# Soft Electrothermal Meta-Actuator for Robust Multifunctional Control

Hanseong Jo,[1] Pavel Shafirin,[1] Christopher Le,[2] Caden Chan,[1] Artur Davoyan[1,*]

*[1] Department of Mechanical and Aerospace Engineering, University of California, Los Angeles*

*[2] Department of Physics and Astronomy, University of California, Los Angeles*

*e-mail: davoyan@seas.ucla.edu*

**Abstract:** Soft electrothermal actuators are of great interest in diverse application domains for their simplicity, compliance, and ease of control. However, the very nature of thermally induced mechanical actuation sets inherent operation constraints: unidirectional motion, environmental sensitivity, and slow response times limited by passive cooling. To overcome these constraints, we propose a meta-actuator architecture, which uses engineered heat transfer in thin films to achieve multifunctional operation. We demonstrate electrically selectable bidirectional motion with large deflection ($\geq$28% of actuator length at 0.75 W), suppressed thermal sensitivity to ambient temperature changes when compared to conventional actuators ($>$100$\times$ lower), and actively forced return to the rest state, which is 10 times faster than that with passive cooling. We further show that our meta-actuator approach enables extended ranges of motions for manipulating complex objects. Versatile soft gripper operations highlight the meta-actuator's potential for soft robotics and devices.

## MAIN TEXT

Soft electrothermal actuators find wide use in diverse fields of science and technology, from soft robotics [1-5] through wearable devices [5-8] to micromanipulation [9-12]. Such actuators are of great interest due to low-voltage actuation, simplicity of design, compliance, thin-film form factor, and complex motion [13-15]. Several notable demonstrations have been shown recently, including, among others, reprogrammable smart systems [16-17] and soft robots that combine action and perception, mimicking biological structures [18-20]. Essentially, electrothermal actuation uses electric power to generate heat, which is then converted to actuation by various types of heat-driven mechanical deformations [21], including strain mismatch structures [22-25], shape memory alloys [26-29], and liquid crystals [30-35]. However, the very mechanism of heat-driven actuation faces a number of fundamental limitations. Firstly, as actuation depends on heating, the system is naturally susceptive to environmental temperature perturbations. Even a moderate rise in ambient temperature could significantly alter actuator performance, therefore limiting the scope of applications. Secondly, once actuated, the original rest state is reached through passive cooling. As a result, actuator's temporal response is limited by passive cooling efficiency. Furthermore, passive cooling becomes inefficient when operating close to ambient temperature. Thirdly, conventional electrothermal actuators move in a single, predetermined direction. Such a limitation in the direction of motion sets strict constraints on the range of geometries where conventional electrothermal actuators can be used. This is where principles of metamaterial design could be used to augment conventional thin-film electrothermal actuators with additional functionalities.

Metamaterials are artificial structures that exhibit functions that are not readily accessible by conventional materials [36-41]. Previously, a range of mechanical metamaterials with emergent properties have been demonstrated, including such systems as ultra-stiffness [42-44], ultra-anisotropicity [45-47], negative Poisson's ratio [48-50], and dynamic tunability [51-52], which conventional materials cannot provide. In the context of flexible, thin-film structures, mechanical metamaterial designs offer functions such as ultra-stretchability [53-56] and dynamic shape morphing [57-61]. Such a metamaterial approach is not limited to passive mechanics of films and can be expanded to engineer active structures integrated with soft actuators [62-63], including complex 3D robotic structures [64-67], adaptive temperature control devices [68-69], and biomedical devices [70-72].

Here, inspired by the principles of metamaterial design, we propose and demonstrate a soft electrothermal *meta-actuator* (Fig. 1a), which exhibits multifunctional responses not accessible by conventional electrothermal actuator designs (Fig. 1b). Through engineering heat transfer in thin films, our meta-actuator allows for an electrically selectable bidirectional motion, is insensitive to environmental thermal perturbations, and exhibits return response faster than regular electrothermal actuators. Specifically, our experiments and field tests reveal that the meta-actuator is independent of environmental perturbations of up to 20°C in temperature and allows more than 10 times faster return to the original rest state. We further show that owing to the bidirectionality of motion, our design allows manipulating objects in difficult-to-reach settings, such as tubes, as well as manipulating complex-shaped objects.

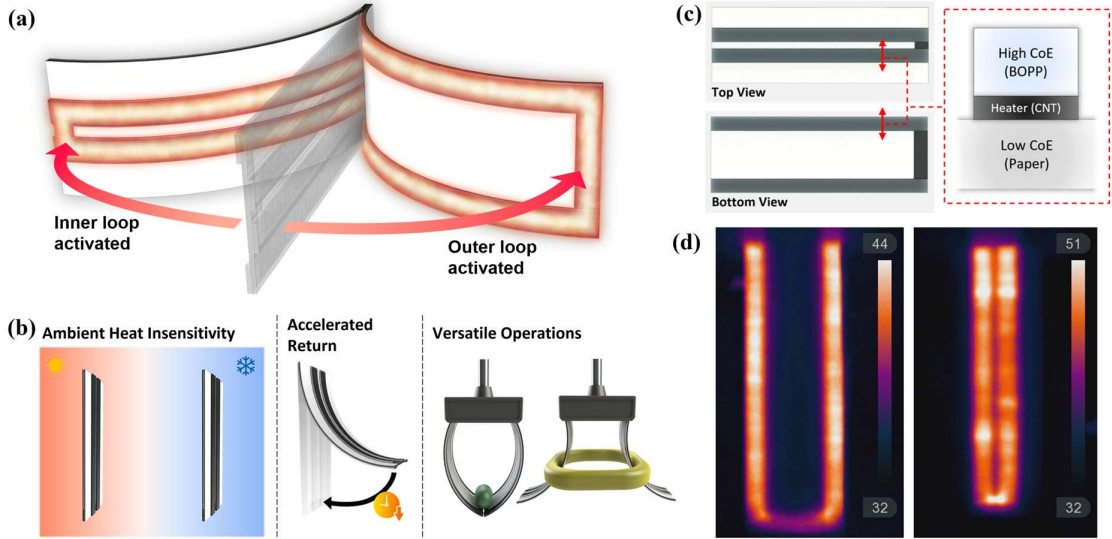

***Figure 1. Electrothermal meta-actuator concept, design, and operation.*** *(**a**) Schematic illustration of a meta-actuator: independent heater loops on opposite sides of a thin film membrane allow for a bidirectional motion and multifunctional operation. (**b**) Conceptual overview of key functions enabled by the meta-actuator, including ambient heat insensitivity, accelerated return, and versatile object manipulation. (**c**) Schematic illustration of top and bottom views of meta-actuator design. Inset shows a cross-sectional view with key materials layers outlined. (**d**) Thermal camera photographs for either outer heater loop (left) or inner heater loop (right) activated. Input power is 0.75 W.*

Figure 1a shows a conceptual illustration of our meta-actuator design. Here, we embed two electric heater loops, one positioned on the top side and the other on the bottom side of a thin-film substrate layer. Our design employs heater loops of different sizes, with a larger loop enclosing the inner one (see Figs. 1a and 1c) with a small gap (~3.3 mm). Such a design ensures minimal thermal cross-talk between the heater loops, which is minimized when convective heat exchange with the ambient exceeds that of heat conduction along the substrate. The latter condition is satisfied when $\frac{hw\Delta d}{kt} \gg 1$, where $h$ is the convective heat transfer coefficient, $w$ is the heater width, $\Delta d$ is the center-to-center distance between two independent heater elements, $t$ is substrate thickness, and $k$ is substrate thermal conductivity. Evidently, thermal cross-talk decreases as the substrate thickness is reduced. As such, the use of thin films with low thermal conductivity allows minimizing spacing between heater elements. For our meta-actuators, we obtain $\frac{hw\Delta d}{kt} \approx 123$ with $h = 10$ W m$^{-2}$ K$^{-1}$, which suggests negligible interference between heater loops and their independent operation for heaters spaced $\geq 1\ mm$ apart from each other (as we show below, such independence is critical for multifunctional operation). In contrast, placing heater loops directly opposite from each other would lead to a strong thermal cross-talk due to out-of-plane heat transfer through the substrate (~6,000 times higher out-of-plane heat transfer as compared to the in-plane one for our structures). We maintain reflection symmetry in the design of heater loops to minimize undesirable twisting motion [73]. Figure 1c provides further details on our meta-actuator design. The actuator dimensions are 35 mm wide and 100 mm long. We use paper with a density of 75 g/m$^2$ (100 μm thickness) as the substrate layer. Paper has a relatively low thermal conductivity of ~ 0.05 W m$^{-1}$ K$^{-1}$ [74], ensuring that in-plane heat transfer is minimized. In addition, while paper exhibits a low coefficient of thermal expansion (CTE) of 4 ppm/K [75], its hygroscopic response leads to net material shrinkage as temperature increases (due to decreasing relative humidity) [76]. Atop the thin film substrate, carbon nanotube heater loops (~ 6.5 mm wide) are deposited by stencil printing and then covered with strips of biaxially oriented polypropylene (BOPP), a material with one of the highest coefficients of thermal expansion (137 ppm/K [77]). See Methods for more details on sample fabrication. Upon passing electrical current, carbon nanotube heaters (resistivity ranges from 1 – 2 k Ω) generate heat. Heating, in turn, leads to the actuator curling caused by the strain mismatch originating from the difference in the coefficients of thermal expansion between paper and BOPP. While here, for the sake of concept demonstration, we choose to operate with commonly available materials, such as paper and BOPP, our design principles can be transferred to other material combinations as well. Hence, by choosing materials with a higher difference in the coefficients of thermal expansion, a larger actuation response is expected. Figure 1d shows thermal camera photographs of the fabricated meta-actuator operated under a 0.75 W power input. Clearly, heat diffusion outside of the deposited carbon nanotube heaters is small, which verifies that the in-plane heat transfer is negligible and ensures independent functioning of the two heater loops.

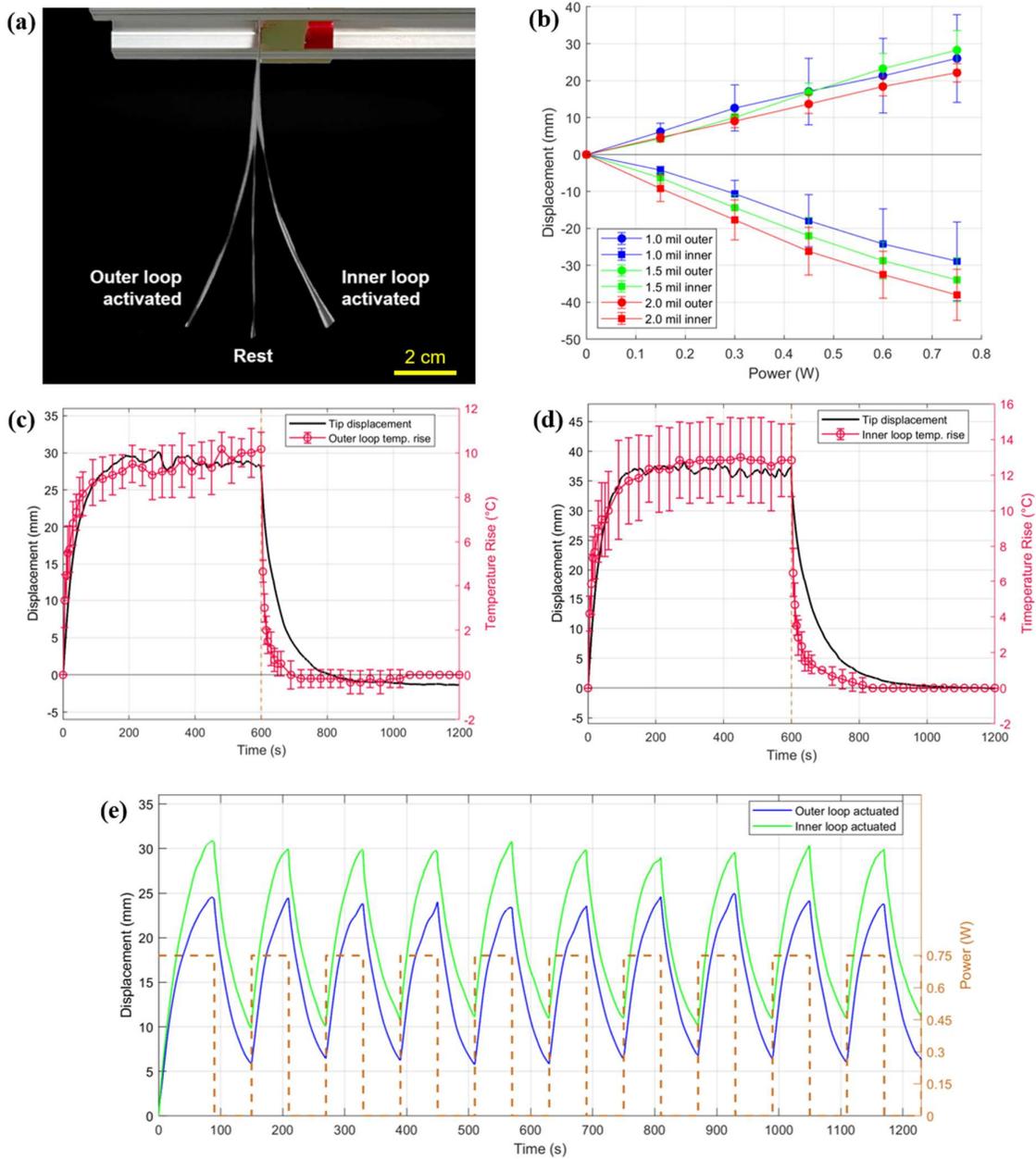

***Figure 2. Meta-actuator deflection and temperature measurements.*** *(**a**) Superimposed photographs showing the actuator in different states: at rest, when the inner heater loop is activated, and when the outer heater loop is activated. (**b**) Tip displacement as a function of input power for actuators with different BOPP layer thicknesses (25 µm, 38 µm, 51 µm). (**c**) and (**d**) Actuator displacement and heater temperature rise as a function of time for outer heater loop (**c**) and inner heater loop (**d**) activation, respectively. (**e**) Cyclic measurement of actuator response for outer and inner heater loop actuation scenarios. Power input in panels (**c-e**) is 0.75 W.*

Moving forward, we study actuator motion and actuation when either the inner or outer heater loops are activated. Figure 2a shows a superimposed image of the actuator operation at rest, with the inner heater loop activated, and the outer heater loop activated, respectively. Evidently, by choosing which of the heaters to activate, the direction of actuator motion can be controlled – a function that is not accessible with conventional electrothermal actuators. We note that several prior works have also explored thin-film actuators that exhibit bidirectional motion. Common strategies involve using multiple types of stimuli [78-81] or using actuation mechanisms that respond to the polarity of the electrical input [82]. Nonetheless, these strategies are complex to implement or place demanding requirements on the environment. Another approach to achieving bidirectional motion makes use of a thermal gradient across a 500 μm thick sandwich structure created by multiple heaters [84]. However, sustaining a high thermal gradient across thin film structures is challenging. To this end, our double-sided thin-film meta-actuator design does not require any special environmental preparations and is fully controlled by electrical stimuli.

In Fig. 2b, we study actuator displacement (measured 1 cm from the actuator tip; see Methods for details) as a function of input electrical power and BOPP thickness for both inner and outer loops activated. As expected, as the input power grows, the actuator deflection from the rest state increases. The dependence is nearly linear up to 0.75 W of applied power for actuators with three different BOPP thicknesses (25 μm, 38 μm, and 51 μm, respectively). Figures 2c and 2d show heater temperature rise, $\Delta T$, and related actuator displacement as a function of time for an actuator with a 51 μm thick BOPP layer, when either its outer or inner heater loop is activated, respectively (here 0.75 W of electric power is applied for 600 s). Initially, when electric power is applied, at first ~50 s, we observe rapid growth in both heater temperature and actuator displacement for both of the activation scenarios. After ~100 s, the temperature approaches a saturation point dictated by a power balance between Joule heating and passive convective cooling. As a result, after some time, actuator displacement also saturates, reaching a maximum value. The inner heater loop exhibits a faster displacement rise time (~150 s) and larger value of maximum displacement (~37 mm, ~132% of the outer loop's) as compared to the outer heater loop (~200 s rise time, ~28 mm maximum displacement (~28% of its length), respectively). The maximum displacement is dictated by the respective heater loop saturation temperature. Hence, the inner loop reaches $\Delta T \sim 12.5°C$, which is ~132% of that for the outer loop. We attribute this difference to the effective area of heater loops. With further optimization of the meta-actuator design, this difference can be mitigated. Upon removing electrical power, the heaters passively cool down due to heat exchange with the ambient environment. While the temperature drop is fast, mechanical response demonstrates a significant lag. The actuator requires ~150 s (after outer loop deactivation) and ~170 s (after inner loop deactivation) to relax to 10% of its saturated maximum displacement and over 400 s to reach <0.1 mm deviation from the rest state. We also observe a slight shift of the rest position for the case of outer heater loop activation ($\simeq -1.3 \; mm$), which we attribute to plastic deformation of the meta-actuator, see also Fig. 2c. To check whether actuator performance is reproducible, we perform cyclic actuation measurements, Fig. 2e. Here for each of the heater loops we turn power (0.75 W) on an off every 60 s. Figure 2e clearly shows that when the outer heater loop is activated, the actuator tip displacement reproducibly cycles between ~6 mm and ~18 mm. For the inner heater loop activation, actuator displacement is cycled between ~11 mm and ~30 mm. In this measurement, as the cycling period is 120 s, the actuator does not reach either full saturation or complete relaxation, which suggests that enhanced control strategies are required for applications demanding rapid motion.

Next, we demonstrate that our meta-actuator design, owing to its double-sided nature, provides stability against ambient temperature variations. When ambient environment temperature grows, a conventional one-sided electrothermal actuator is uniformly heated. This heating triggers the thermal expansion of materials and causes curling, an effect nearly identical to actuation with an applied electric current. As such, conventional electrothermal actuators are limited to a relatively narrow range of operating temperatures. In contrast, for our meta-actuators, ambient heating of the device activates thermal loops on both sides of the actuator, which, in turn, negate each other in the presence of ambient temperature variations (Fig. 3a). To experimentally verify the insensitivity of the meta-actuator to ambient temperature variations, we measure the curvature of both our double-sided meta-actuator and conventional single-sided actuator at their respective rest states (i.e., with no electric current applied) as a function of the ambient air temperature (see Methods). Figure 3b plots curvature for two different actuator types as a function of temperature rise, $\Delta T$, as the ambient air temperature is varied from room temperature (~23º C) to ~43º C. Measurements clearly show that the ambient air temperature greatly influences the rest state of the conventional single-sided actuators. At the same time, the meta-actuators, owing to their double-sided design, show minimal variation with respect to their rest state. For conventional single-sided actuators, actuator curvature increases approximately linearly with temperature at small $\Delta T$: $\kappa = c\Delta T$, where $c$ plays a role in thermal sensitivity coefficient. We estimate that conventional actuators exhibit thermal sensitivity of $c_{\mathrm{conv}} \simeq 0.047$ cm$^{-1}$K$^{-1}$ for $\Delta T = [0℃, 10℃]$. In particular, for a conventional single-sided actuator, the temperature rise of just $\Delta T = 5°C$ induces a curvature of $0.29$ cm$^{-1}$ (corresponding effective radius $R = 3.45$ cm), rendering it nearly unusable given its 10 cm length. In contrast, the meta-actuators exhibit negligible curvature changes under the same thermal loading. Across the studied temperature range, meta-actuators show >100× lower overall thermal sensitivity, $c_{\text{meta}} \simeq 0.00041$ cm$^{-1}$K$^{-1}$. We expect that this minimal curvature deviance can be further reduced by optimizing actuator design and parameters.

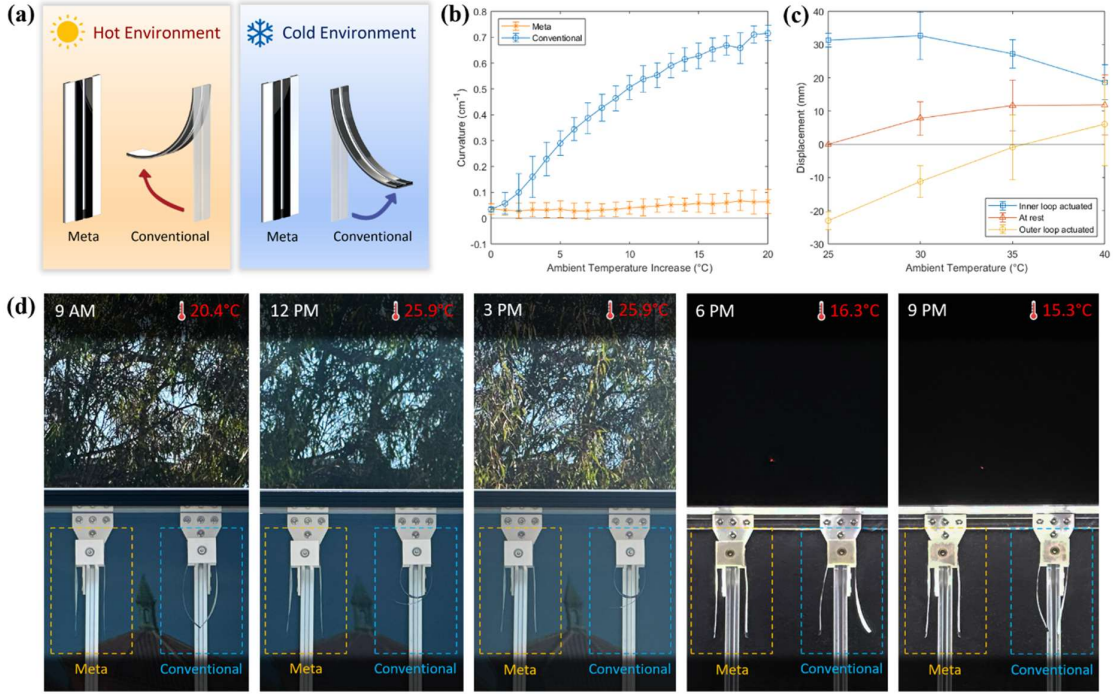

***Figure 3. Ambient thermal insensitivity of meta-actuator.** **(a)** Illustration comparing meta-actuator and conventional actuator response to ambient temperature changes. **(b)** Actuator curvatures as a function of ambient air temperature rise for meta-actuators and conventional actuators, respectively. **(c)** Meta-actuator tip displacement versus ambient temperature at rest and during inner or outer loop activation, respectively (0.75 W input power, displacement measured after 300 s). **(d)** Field test performance comparison for grippers fabricated with meta-actuators and conventional single-sided actuators.*

To verify the meta-actuator performance at elevated temperatures, in Fig. 3c, we measure the meta-actuator tip displacement as a function of ambient air temperature for 3 states: at rest, inner loop activated, and outer loop activated (0.75 W power, measured after 300 s of activation). Across the entire temperature range, actuation in both directions of motion is observed. However, we observe that displacement magnitude decreases with temperature increase. The maximum actuation range drops from 56 mm at $T = 25°C$ to ~13 mm at 40ºC. Such reduction in performance can be attributed to the nonlinearity of the material properties [84-87]. Additionally, we observe that the rest position itself is drifting with temperature. However, this rest position drift shows significant sample-to-sample variability, suggesting that the effect can be minimized through optimization in design and fabrication. To further verify the insensitivity of the meta-actuator design to environmental perturbations, we conducted a proof-of-concept field test. In Fig. 3d, we compare the behavior of two types of grippers: one made with meta-actuators and the other with conventional single-sided actuators. Specifically, we placed fabricated grippers outdoors, where they were subjected to natural daily air temperature variations in Los Angeles. The ambient air temperature first increased to ~26ºC at 12 pm and then fell to ~15ºC at 9 pm. Respective photographic images taken every 3 hours indicate that the meta-actuator gripper maintains its original shape, whereas the grippers made of conventional one-sided actuators exhibit substantial instability. This simple demonstration shows the promise of our meta-actuator design for robust operations against possible ambient temperature fluctuations in field applications.

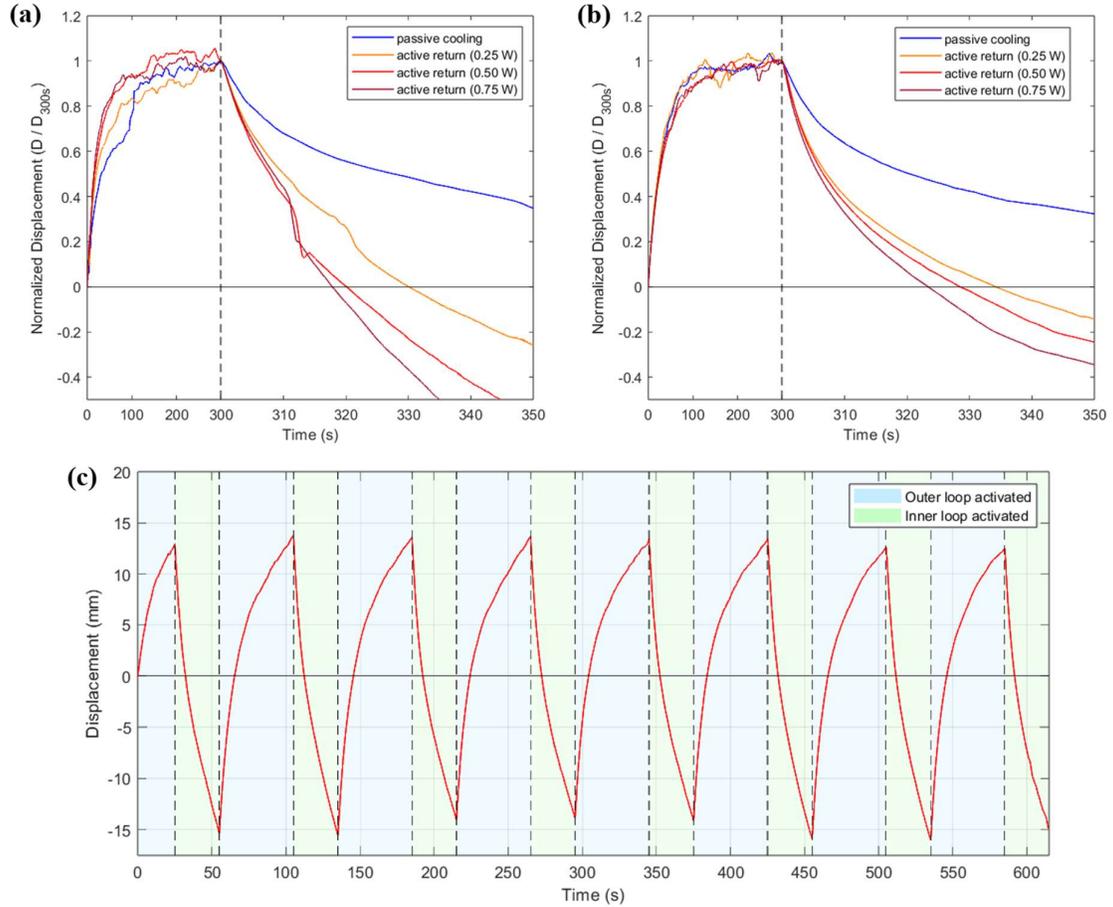

***Figure 4. Meta-actuator forced return and motion control.*** *(**a**) Normalized displacement as a function of time for the inner-loop forced return at different power levels (passive, 0.25 W, 0.50 W, 0.75 W). (**b**) Same as in panel (a) but for the outer-loop forced return. In panels (a) and (b), the timescale is compressed prior to t=300 s for convenience. (**c**) Displacement as a function of time for alternating cyclic actuation of outer (50 s) and inner (30 s) heater loops. The activation power is 0.75W for both the inner and outer heater loops.*

We proceed with a study of active control of meta-actuator motion. In the case of conventional electrothermal actuators, once actuated, the original rest state is reached after passive cooling through heat dissipation to the surrounding environment. As was shown in Fig. 2b and 2c, the passive cool-down phase is rather slow ($\sim$ 400 s to reach the final rest state for our structures). Additionally, in case of plastic deformations or undesirable changes in the material properties, conventional actuators may never completely return to their original rest state (see Fig. 2c). In contrast, our meta-actuator design, owing to its double-sided nature, allows forced return to the rest state. To examine active control using meta-actuator motion, we study meta-actuator displacement as a function of time for several different input power levels. For this purpose, we first activate one of the heater loops for 300 s at 0.75 W; we then remove power from the said heater loop while simultaneously turning on the opposite heater loop. The resulting temporal dynamics for the two possible scenarios are illustrated in Figs. 4a and 4b, respectively (here normalized displacement, defined as a ratio of the displacement at time $t$ to the one at $t = 300$ s is plotted; see also Methods). As expected, supplying higher power to the returning heater loop results in a faster temperature rise and, consequently, a more rapid return to the rest state. For both scenarios, even at low powers of 0.25 W, a dramatic improvement in response time is observed. The actuator reaches the original rest state after 30 s for the inner-loop forced return and after 34 s for the outer-loop forced return, respectively. For comparison, in 30 s, a passively cooled actuator still exhibits >0.4 normalized displacement. Notably, for the inner-loop forced return, the actuators have shown snap-through buckling effects for all power levels, which is attributed to the geometric configuration and the counteracting stress profiles of the two loops. Utilizing the accelerated return mechanism enables fast bidirectional cyclic motion, as shown in Supporting Video S1 and Fig. 4c. Figure 4c plots the actuator displacement during alternating activation of the inner (30 s) and outer (50 s) heater loops at 0.75 W. Repeatability of fast bidirectional motion is clearly evident.

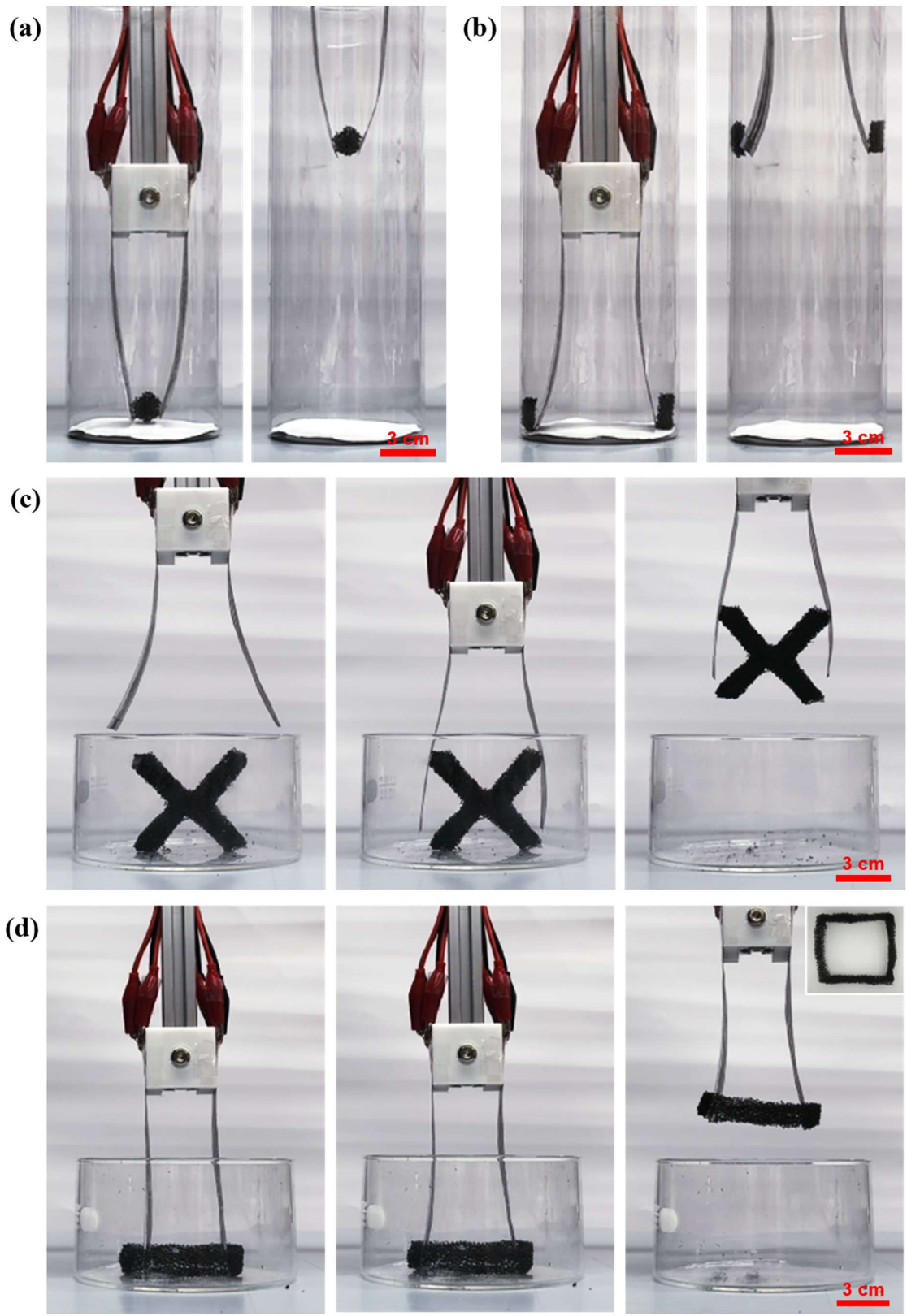

*Figure 5. Complex objects manipulation capabilities with meta-actuator soft grippers.* *(a)* *Gripping and transporting of an object smaller than gripper jaw at rest.* *(b)* *Object retrieval*

*within a confined tube via gripper jaw opening and wall pressing. **(c)** Handling of oversized objects via gripper jaw opening. Gripping and transportation of an X-shaped object larger than the original gripper jaw at rest is shown. **(d)** Handling of large hollow objects by jaw expansion within the opening. Gripping and transportation of a rectangular ring-shaped object are shown. Inset: top view of the rectangular ring-shaped object.*

Finally, we explore the utility of our multifunctional meta-actuators for object manipulation. In scenarios employing conventional electrothermal actuators, the inherent unidirectionality limits device versatility and functionality. In contrast, our meta-actuator selective and bidirectional motion introduces flexibility and extra degrees of freedom for object manipulation. For demonstration purposes, we study object manipulation with a 2-fingered soft gripper made of meta-actuators. Unlike conventional electrothermal grippers, such a "meta-gripper" supports both opening and closing motions. As a result, such meta-grippers could perform more complex manipulations, inaccessible to conventional electrothermal actuator grippers. In Fig. 5a and Supporting Video S2 we show the gripper (with an initial jaw opening of 40 mm) transporting a spherical object (diameter: 16 mm) made of polyurethane foam. This type of simple gripping operation can also be performed by conventional electrothermal grippers. However, the following scenarios highlight the meta-gripper's unique capabilities beyond those of conventional grippers. First, retrieving objects situated near the wall within a narrow tube via conventional jaw closure is often infeasible. To tackle this scenario, the meta-gripper utilizes its opening capability: it expands its jaws to press the object against the tube wall before dragging it out. This procedure is shown in Fig. 5b and Supporting Video S3, with 8 mm in wide cuboid foam blocks in an 80 mm diameter transparent tube. In addition, the meta-gripper can actively widen its jaw prior to grasping, enabling the transport of objects larger than its original jaw opening. This capability is shown in Fig. 5c and Supporting Video S4, where the meta-gripper opens its jaw from 40 mm to 92 mm to securely hold and transport an "X"-shaped object with a width of 65 mm, i.e., more than 160% of the gripper's original jaw. Moreover, the meta-gripper is capable of handling large hollow objects with openings (e.g., toroids): it first slides into the target object's opening and then expands its jaw to grab and lift. Figure 5d and Supporting Video S5 demonstrate this capability using a rectangular ring-shaped foam (outer width: 70 mm, opening width: 54 mm). Although we focus here on the utility of the meta-actuators in a simple example of a 2-fingered gripper, we anticipate that meta-actuators can be used in more complex gripper configurations to enable even more versatile object manipulation.

In summary, we have proposed and demonstrated a soft electrothermal meta-actuator architecture that enables new functions that are not accessible to conventional electrothermal actuators. Specifically, we showed that our actuators exhibit bidirectional motion, allow for >100x better stability against ambient temperature perturbations, and enable forced returns to the original position that is >10x faster, as compared to passive cooling. We further demonstrated that our meta-actuators allow for new modalities in gripping and object manipulation. As such, our work lays the foundation for the use of metamaterial design principles to design versatile and multifunctional systems for future adaptive soft robotics. We anticipate that with further optimization of materials and designs, meta-actuator capabilities and functionalities can be further enhanced to meet even more demanding application requirements.

Experimental Section/Methods

***Fabrication of actuators:***

Fabrication of actuators follows a modified stencil-printing technique adapted from previous work [76]. First, Kapton tape (Bertech, USA), comprising a 51 µm film layer and a 38 µm adhesive layer, serves as the masking material. Using this mask, a water-soluble multi-walled carbon nanotube (MWCNT) paste (NovaCP-CNT-A2, Novarials, USA) containing approximately 5 wt% MWCNTs (~10 nm diameter) is applied onto a standard copy paper substrate (75 g/m²) via squeegee blade. Once the CNT heater pattern is air-dried, the process is repeated on the opposite side for meta-actuator fabrication; this second printing step is omitted for unidirectional actuators (shown in Fig. 3). Then, strips of biaxially oriented polypropylene (BOPP) tapes (3M, USA) are cut and applied onto the CNT heaters. Three different BOPP tapes are used, featuring distinct BOPP layer thicknesses but sharing a uniform adhesive thickness of 16 µm: 371 (25 µm BOPP), 373 (38 µm BOPP), and 375 (51 µm BOPP). Finally, copper electrodes are attached to complete the fabrication process.

***Characterization and Measurements:***

Actuator displacement measurements were conducted with a laser displacement sensor (optoNCDT 1320, Micro-Epsilon, USA) mounted on a vertical linear rail. The specific measurement strategy depended on the experiment. For power-response characterization (Figs. 2b, 3c), the displacement at the actuator tip was recorded. For time-response measurements presented in Figs. 2c, 2d, and 2e, the sensor instead tracked a fixed reference point marked on the actuator (1 cm from the tip along the actuator's length) for stable real-time tracking. For characterizing the active return dynamics (Figs. 4a, 4b, 4c), the sensor's position was fixed (75 mm vertically below the actuator's highest point) to ensure continuous tracking. Temperature measurements were obtained using a thermal camera (ShotPRO, Seek, USA) by averaging readings from six different points (¼, ½, and ¾ points along the 2 long edges of the heater). Curvature measurement was performed with a digital camera and image processing algorithms. Here, curvature, $\kappa$, is defined as $\kappa = \frac{1}{R}$, where $R$ is the effective radius actuator. Points at ¼, ½, and ¾ along the actuator's length within the captured images, using these positions to define a unique circle. We then used this circle's radius ($R$) to determine the curvature ($\kappa$). For controlling ambient air temperature (Figs. 3b, 3c), an enclosure was constructed from acrylic plates. Inside this enclosure, a heated metal plate controlled by a hot plate regulates the temperature. Actuators were shielded from direct radiation from the heated metal plate, and 3 sensors placed near the actuator monitored the air temperature. For the field test shown in Fig. 3d, the actuators were placed in a bottomless semi-enclosure, allowing for natural airflow, and opaque acrylic plates formed the sides to block direct solar radiation.

**Supporting Information**

Supporting Information is available from the Wiley Online Library or from the author.

Acknowledgements

Authors thank useful discussions with Mozakkar Hossain and Tom Joly-Jehenne. Research is supported by NASA Grant# 80NSSC23K0076.